\documentclass[letterpaper, 10 pt, conference]{ieeeconf}  

\IEEEoverridecommandlockouts                              

\usepackage{siunitx} 
\usepackage{soul}
\usepackage{xcolor}
\usepackage{graphicx}
\sethlcolor{yellow}
\usepackage{cite}
\usepackage{booktabs}
\usepackage{censor}
\usepackage{url}
\usepackage{hyperref}
\usepackage{float}

\title{\LARGE \bf
GlassFormer: Learning Real-time Glass Segmentation using Radar-Depth Fusion
\thanks{Code available at: \url{https://github.com/Suhani92/GlassFormer}}
}
\author{Suhani Grover, Astik Srivastava, Viswas Dinesh, Avinash Sharma, and Madhava Krishna
\thanks{Suhani Grover, Astik Srivastava, Viswas Dinesh and Madhava Krishna are with Robotics Research Center, IIIT Hyderabad, India. Email: suhani1077@gmail.com, (astik.srivastava, viswas.dinesh)@research.iiit.ac.in, mkrishna@iiit.ac.in}
\thanks{Avinash Sharma is with IIT-Jodhpur, India. Email: avinashsharma@iitj.ac.in }
}

\begin{document}

\maketitle
\thispagestyle{empty}
\pagestyle{empty}

\begin{abstract}
Transparent surfaces are ubiquitous in built environments, yet they remain a persistent failure case for robotic perception. RGB cameras perceive the background behind glass rather than the surface itself, while depth sensors such as LiDAR, time-of-flight, and RGB-D often return invalid or background measurements in transparent regions. As a result, systems that rely solely on optical sensing may misinterpret glass walls, doors, or mirrors as free space, compromising safe and reliable navigation.

Existing glass segmentation approaches address this by learning visual cues such as reflections, boundaries, and semantic context from RGB images. While effective under favourable lighting and viewing conditions, these cues degrade in low-light environments, under glare, or when glass surfaces are featureless or partially occluded. In this work, we propose a multimodal framework that fuses
millimetre-wave radar with RGB-D sensing for real-time transparent surface segmentation.
Radar reflects strongly off glass surfaces, providing a geometric cue that
remains reliable precisely where vision and depth fail. We exploit this cross-modal inconsistency to generate a radar-guided spatial prior, which is integrated into a lightweight transformer-based segmentation network, GlassFormer, via cross-modal attention. 

To evaluate our approach, we collect a synchronized RGB-D-radar dataset spanning diverse glass types, including doors, windows, and mirrors, across a wide range of lighting conditions from daylight to near-dark environments. We report results on a mixed-condition test split covering all scene types and a dedicated low-light split designed to stress vision-only methods. GlassFormer achieves 0.88 mIoU on the mixed split, and 0.59 mIoU on the low light split, demonstrating substantial robustness gains over vision-only baselines while maintaining real-time performance on resource-constrained platforms.

\end{abstract}

\section{Introduction}
\label{sec:intro}

Transparent surfaces such as glass doors, partitions, and windows remain a fundamental failure case for robotic perception systems. Unlike most obstacles, glass is simultaneously invisible to conventional depth sensors and visually ambiguous to RGB-based methods, making it difficult to detect reliably across real-world operating conditions.

This challenge stems from the two complementary properties of glass. First, glass lacks intrinsic colour or texture; its visual appearance is determined almost entirely by the scene around and behind it, making it ambiguous to RGB-based methods, particularly under low lighting or featureless backgrounds. Second, glass simultaneously reflects, refracts, and transmits incident electromagnetic energy. As a result, RGB-D cameras, time-of-flight sensors, and LiDAR frequently misinterpret glass as free space or return invalid depth at transparent interfaces, which is a drawback that is intrinsic to the sensing modality and persists irrespective of ambient illumination. RGB-only methods that rely on visual cues such as reflections and boundaries degrade specifically under low-light, glare or featureless conditions as these cues become unreliable when insufficient texture or lighting is available. While ultrasonic sensors can detect transparent obstacles at short range, their sparse and limited spatial measurements make dense segmentation challenging. Prior work has explored a range of visual cues to address this, including boundaries, reflections, polarization and semantic context ~\cite{enhancedboundary,richcontext,glasssemnet,polarizationdiff,multiscales,11123849}. While these methods perform well under favourable conditions, their core assumptions - visible boundaries, identifiable reflections and adequate illumination break down in scenarios relevant to practical deployment, such as low-light corridors, high-glare atriums, and featureless glass panes. Large-scale datasets such as ClearPose~\cite{clearpose} and TRansPose~\cite{transpose} have significantly advanced transparent object perception. However, they primarily target object-level manipulation or multispectral perception rather than dense scene-level glass segmentation for mobile robotics. This reflects a fundamental limitation of optical sensing in the presence of transparent surfaces. 
\begin{figure}[t]
  \centering  \includegraphics[width=\columnwidth]{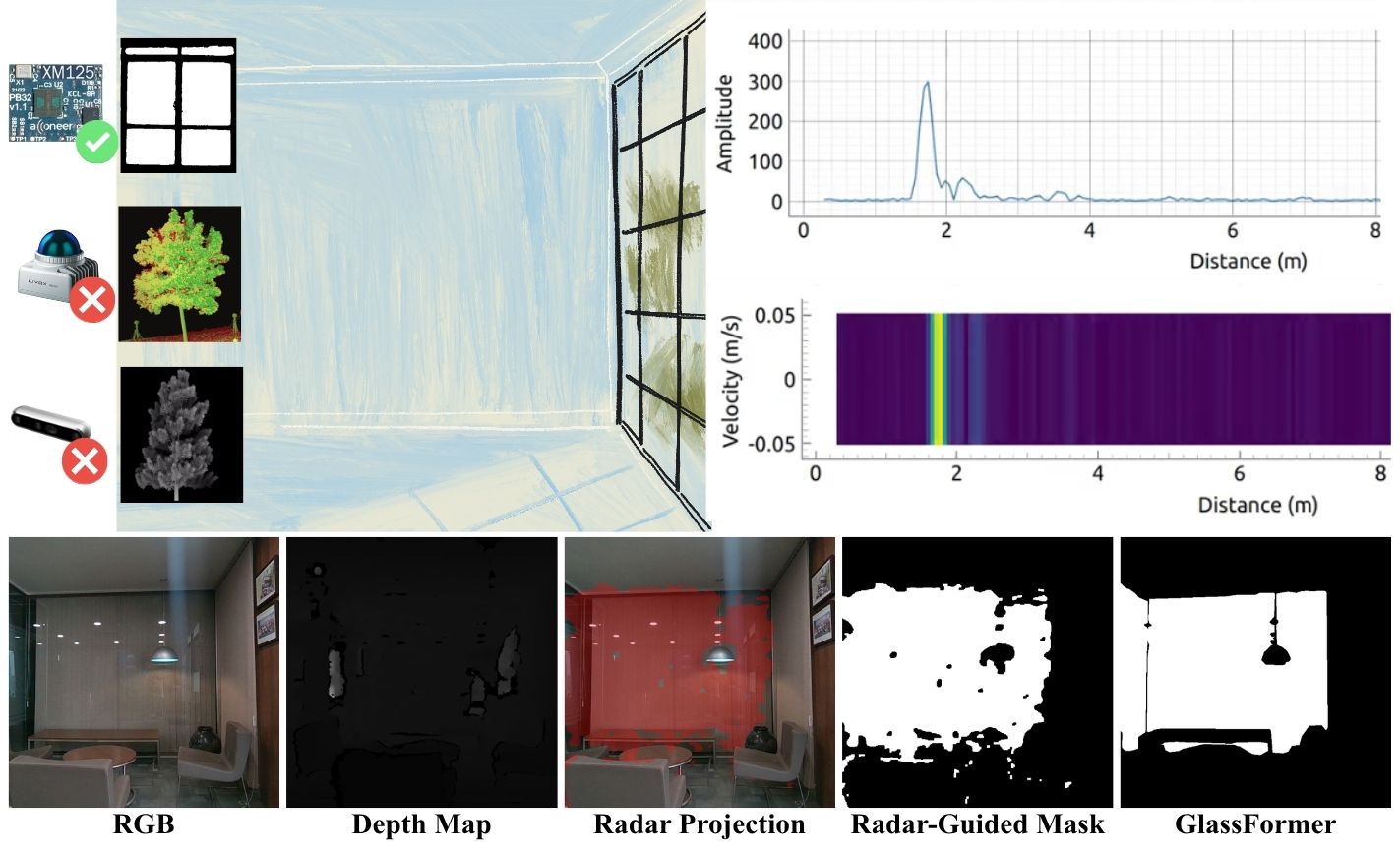}
  \caption{RGB-D and LiDAR fail on glass, returning background depth or invalid returns through the surface, while mmWave radar reflects strongly off it. This cross-modal disagreement yields a radar-guided prior (pink) over likely transparent regions, which GlassFormer fuses with RGB to segment glass reliably, even in low light.}
  \label{fig:range_profile}
\end{figure}
We observe that the millimeter-wave (mmWave) radar occupies a complementary position in this sensing landscape. At 60\,GHz, radar wavelengths are insensitive to ambient lighting and crucially, are not refracted through glass, addressing illumination-dependent failure mode of RGB cues and the illumination-independent failure mode of depth sensing simultaneously. While radar's interaction differs physically for transmissive glass versus specular mirrors, our evaluation and qualitative results (Fig. ~\ref{fig:sota_qualitative}) indicate the method performs effectively on both surface types. Despite widespread adoption of radar in robotics for proximity sensing and automotive perception, its use as a spatial prior for vision-based segmentation has not yet been explored. While radar alone cannot produce pixel-dense segmentation masks, even a coarse range estimate can narrow down the search space considerably for a downstream vision model.

In this work, we introduce GlassFormer, a radar-guided transparent surface segmentation framework designed for real-time deployment on resource-constrained robotics platforms. 
A mmWave pulsed coherent radar estimates the range of dominant scene reflectors, which is projected into the image plane and fused with RGB-D depth inconsistency cues to generate a coarse spatial prior localizing transparent surface candidates. The prior is then injected into a SegFormer-B2 backbone using a cross-modal RadarAttention module, which modulates deep semantic features with range-conditioned spatial context.

Our main contributions are as follows:
\begin{itemize}
    \item A radar-guided mask generation pipeline that fuses millimeter-wave range measurements with RGB-D depth inconsistencies to produce coarse but reliable spatial priors for transparent surface localization in the image plane.
    \item GlassFormer, which injects this prior into a SegFormer-B2 backbone via cross-modal RadarAttention, achieving substantial gains over state-of-the-art methods under low-light and high-glare conditions where vision-only approaches degrade.
    \item Synchronized RGB-D-radar dataset for transparent surface segmentation comprising annotated frames across diverse scenes, including low-light and high-glare conditions underrepresented in existing benchmarks.
    \item An open-source ROS2 driver enabling real-time integration of pulsed coherent radar within robotic perception pipelines.
\end{itemize}

\section{Related Work}
\label{sec:related}
Reliable perception of transparent surfaces is critical for safe robotic operation. This section summarizes prior work on transparent surface detection, including vision-based segmentation methods and multimodal sensing approaches that combine complementary sensors.

\subsection{Vision-Based Transparent Surface Segmentation}
Early work on transparent surface segmentation works were done using only RGB images. 
~\cite{gdnet} were the first to introduce a dedicated glass detection
network, using large-field contextual feature integration to capture
global cues.
Although these methods give good results, RGB cannot provide good cues for glass objects. Subsequent methods incorporated glass-specific physical priors to address
this. ~\cite{richcontext} exploit reflections as a refinement
signal, while ~\cite{enhancedboundary} focus on boundary
learning via a refined differential module and edge-aware graph
convolution. GlassSemNet~\cite{glasssemnet} takes a different approach,
reasoning over semantic co-occurrence patterns to provide contextual
priors through a dual-backbone architecture combining SegFormer with a
semantic ResNet branch. Progressive Glass Segmentation~\cite{Yu_2022} further improves segmentation through coarse-to-fine refinement, while PanoGlassNet~\cite{10504145} exploits panoramic RGB and intensity images for multimodal glass detection in large-scale environments.

TransLab~\cite{translab} encodes boundary
information directly into the transformer pipeline, and Trans4Trans~\cite{trans4trans} targets real-world navigation assistance for visually impaired users. 
These models still assume good lighting and weak specular reflection. 

\subsection{Multimodal Perception for Transparent Surfaces}
The limitations of optical sensing have motivated a range of multimodal
approaches. RGB-D cameras are the most common pairing: several works fuse depth with RGB features via cross-modal attention~\cite{dagnet,rgbdglass, learningdepthestimationtransparent}
or weighted feature fusion~\cite{weightedff} to exploit the inconsistency between depth failures and visual context. However, these methods rely on depth, which fails on transparent surfaces.

To overcome this, researchers have turned to sensors whose physical interaction with glass differs fundamentally from optical devices.
LiDAR-based approaches exploit intensity variance ~\cite{2dlidar,lidaroccgrid} and multi-echo processing ~\cite{5409636,KOCH2017296} but are expensive and typically limited to 2D. Ultrasonic sensors can confirm transparent surfaces at
short ranges and have been fused with RGB-D cameras for object
reconstruction~\cite{ultrasonic_rgbd}, but they have a very small spatial resolution and are extremely sensitive to noise. 
Polarization cameras capture the rotation of light waves off specular surfaces and have been combined with both deep
learning~\cite{polarizationdiff} and LiDAR~\cite{lidar_polarisation}
for glass detection, but require expensive and specialised hardware that make them impractical for robotic applications.
Thermal imaging provides illumination-invariant cues and has been fused
with RGB for segmentation~\cite{rgbthermal_glass}.

Most closely related to our work is the radar-RGB-D fusion framework of
FuseGrasp~\cite{fusegrasp}, which uses radar sensing to assist robotic grasping of transparent objects. However, FuseGrasp relies on higher-dimensional radar processing and is designed for manipulation rather than real-time scene understanding. In contrast, our approach uses a 1D radar signal to derive a region prior that can be fused with RGB observations for transparent surface segmentation, enabling real-time operation on low-compute platforms. ~\cite{tof_ultrasonic} designed a ToF-ultrasonic fusion system, which demonstrates real-time
transparent obstacle mapping on a low-SWaP aerial platform using only
CPU inference, but this functions for a very limited angular range. These works establish that non-optical range sensors provide complementary and reliable cues where traditional vision-based methods fail.

\section{Preliminaries}
\label{sec:preliminary}

\subsection{Pulsed Coherent Radar and Signal Model}
We use a \SI{60.5}{\giga\hertz} millimeter-wave pulsed coherent radar to detect transparent surfaces that are often invisible to optical depth sensors. The radar estimates target range by transmitting short electromagnetic pulses and measuring their round-trip propagation delay $t_{\mathrm{delay}}$ as:
\begin{equation}
d = \frac{v\, t_{\mathrm{delay}}}{2}, 
\qquad 
v = \frac{c_0}{\sqrt{\varepsilon_r}},
\end{equation}
here, $v$ is the wave speed in the medium, $c_0$ is the speed of light in vacuum, and $\varepsilon_r$ is the relative permittivity of the medium. The factor of 2 accounts for the round-trip travel of the pulse.

For each frame, the radar outputs complex in-phase and quadrature (IQ) measurements as:
\begin{equation}
x[s,d] = I[s,d] + jQ[s,d],
\end{equation}
where $s$ indexes temporal sweeps and $d$ indexes discrete range bins along the radar line of sight. Each signal encodes a magnitude that indicates reflector strength and phase information that captures sub-wavelength displacement. In our work, we use the magnitude profile across range bins to localize dominant reflecting surfaces. 
\begin{figure*}[h]
  \centering
  \includegraphics[width=\textwidth]{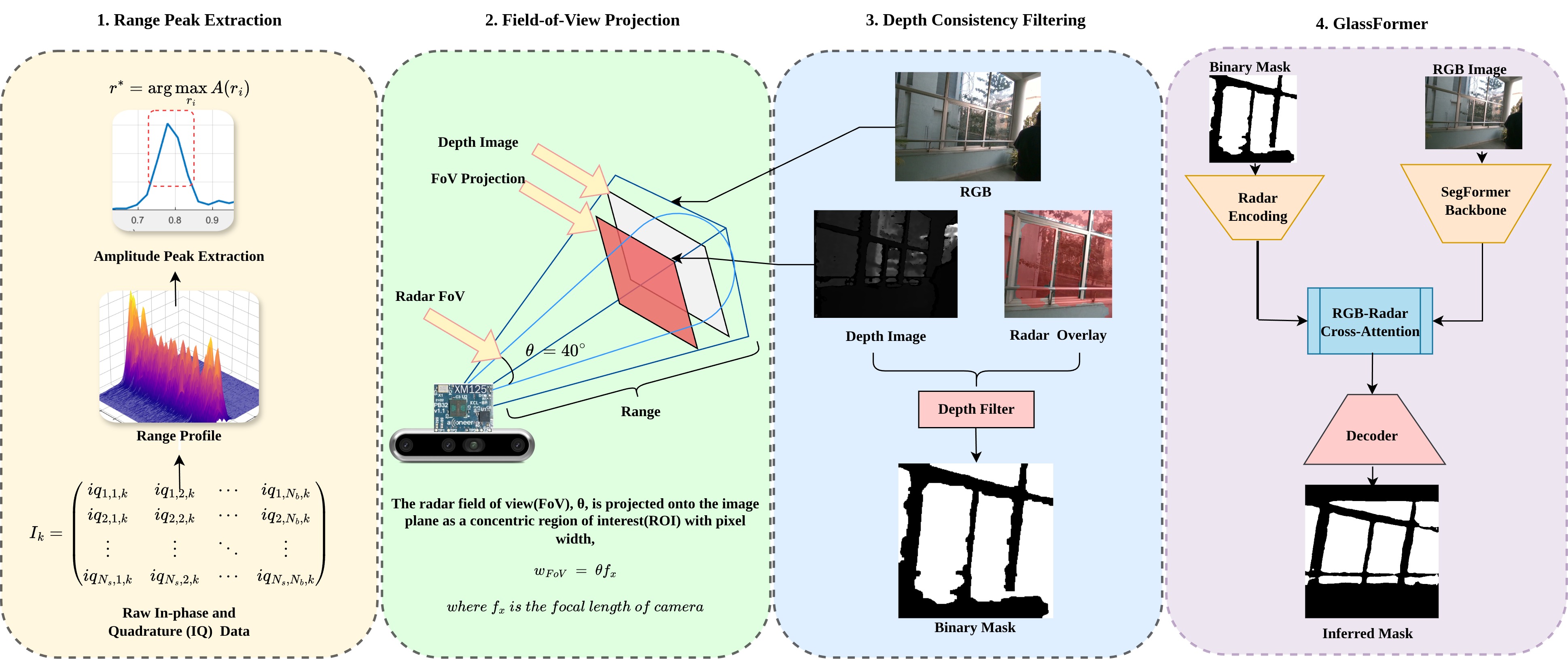}
  \caption{Radar-guided mask generation pipeline. The radar produces a one-dimensional range magnitude profile from which the dominant reflection peak is
extracted to estimate the radial distance dr of the nearest surface. This distance corresponds to the most prominent reflector along the radar boresight. The
radar field-of-view (FoV) is then projected into the RGB image plane using camera intrinsics, defining a region of interest centered at the principal point.
Within this projected FoV, the radar-derived range is compared against per-pixel depth measurements from the RGB-D sensor. Pixels with invalid depth
values or depths significantly larger than the radar-confirmed surface are labeled as transparent candidates, since the depth sensor observes background
geometry through the surface while the radar detects the physical interface. The resulting binary mask serves as a radar-derived transparency prior that
highlights regions where optical sensing is unreliable.}
  \label{fig:radar_mask_generation}
\end{figure*}
Although glass is optically transparent, it exhibits a relative permittivity of $\varepsilon_r \approx 6$--$8$ at millimeter-wave frequencies. The resulting impedance discontinuity at the air-glass interface produces a measurable radar return even when visible and infrared light is transmitted or specularly reflected. Consequently, planar glass surfaces generate consistent amplitude peaks in the radar range profile, whereas structured-light and stereo depth sensors often give invalid or inconsistent measurements at glass boundaries.

\subsection{Sensor Geometry and Configuration}
The radar outputs a one-dimensional range profile per sweep, consisting of discrete range bins corresponding to the reflected amplitude and phase from a specific radial distance. The spacing between bins is determined by the configured step length. In our configuration, 200 range bins are sampled, defining an effective sensing window of approximately \SIrange{0.30}{13.0}{\metre}. 
To improve measurement stability, hardware averaging per sample (\texttt{HWAAS} = 64) is applied, meaning each range bin value is computed from 64 internally averaged pulse measurements to increase signal-to-noise ratio. Additionally, three sweeps are aggregated per frame (\texttt{sweeps\_per\_frame} = 3), providing short temporal averaging while preserving responsiveness for real-time perception. 

\begin{table}[t]
\centering
\caption{Radar configuration for data acquisition}
\label{tab:radar_config}
\begin{tabular}{l c}
\toprule
\textbf{Parameter} & \textbf{Value} \\
\midrule
Start point & 120 \\
Number of range points & 200 \\
Step length & 24 \\
Sweeps per frame & 3 \\
HWAAS & 64 \\
Receiver gain & 13 \\
Pulse Repetition Frequency (PRF) & 8.7 MHz \\
\bottomrule
\end{tabular}
\end{table}
The radar integrates an Antenna-in-Package (AiP) with an approximate half-power beamwidth (HPBW) of $40^\circ$. The beamwidth defines a conical sensing volume centered along the radar boresight within which reflected signals are received with at least half of the peak radiated power. This determines the spatial region projected into the image plane for radar-guided mask generation (Section ~\ref{sec:radar_mask}).

\section{Methodology}

\subsection{Radar-Guided Mask Generation}
\label{sec:radar_mask}

\subsubsection{Range Peak Extraction} 
For each synchronized frame, the radar produces a one-dimensional magnitude profile across discrete range bins (as shown in Fig. \ref{fig:radar_mask_generation}). The dominant reflector is estimated by selecting the bin with maximum intensity:
\begin{equation}
r^{*} = \arg\max_{r_i} A(r_i)
\end{equation}
The corresponding radial distance $d_r$ represents the most prominent surface detected along the radar boresight. This peak-selection approach assumes a dominant single-interface reflection; for thicker glass or multipath conditions, distinct front and back surface returns may occur, in which case the selected peak may not correspond exactly to the near air-glass interface.

\subsubsection{Field-of-View Projection}

Since the radar provides no angular resolution, spatial localization is performed by projecting the conical radar field of view (FoV) into the RGB image plane. 
Given the camera focal length $f_x$ and principal point $(c_x, c_y)$ from intrinsic calibration, the radar FoV $\theta$ is converted to a pixel width:
\begin{equation}
w_{\mathrm{FoV}} = \theta f_x
\end{equation}

This defines a region of interest (ROI) centered at $(c_x, c_y)$ within the RGB image (Fig.~\ref{fig:radar_mask_generation}). Although the radar beam has a conical radiation pattern, we approximate its projection as a rectangular ROI in the image plane for computational simplicity and alignment with pixel grid discretization. All subsequent depth-based reasoning is restricted to this projected FoV. Unlike approaches that assume a frontal robot pose relative to the glass surface, our radar-guided prior is not restricted to perpendicular viewing as the projection remains valid across the full calibrated field-of-view shared between the radar and RGB-D sensor. 

\subsubsection{Depth Consistency Filtering}

Within the projected FoV, a pixel $(x,y)$ is labeled as a transparent candidate if the depth sensor does not corroborate the surface detected by the radar:
\begin{equation}
    M(x,y) = \mathbf{1}\bigl[D(x,y) = invalid \;\lor\; D(x,y) - d_r > \tau\bigr],
    \label{eq:mask}
\end{equation}
where $D(x,y)$ denotes the depth measurement from the RGB-D sensor at pixel $(x,y)$, $d_r$ denotes the dominant radar range obtained via peak extraction from the one-dimensional range profile, and $\tau$ is a positive tolerance margin.
The first condition captures pixels with invalid or missing depth returns. The second identifies pixels whose measured depth lies sufficiently beyond the radar-confirmed surface, indicating a likely transparent interface between the sensors and background geometry. The tolerance $\tau$ serves as a practical cross-modal gating parameter to accommodate residual calibration error, surface thickness, and multi-path effects.
\begin{figure*}[t]
  \centering  \includegraphics[width=1.0\linewidth]{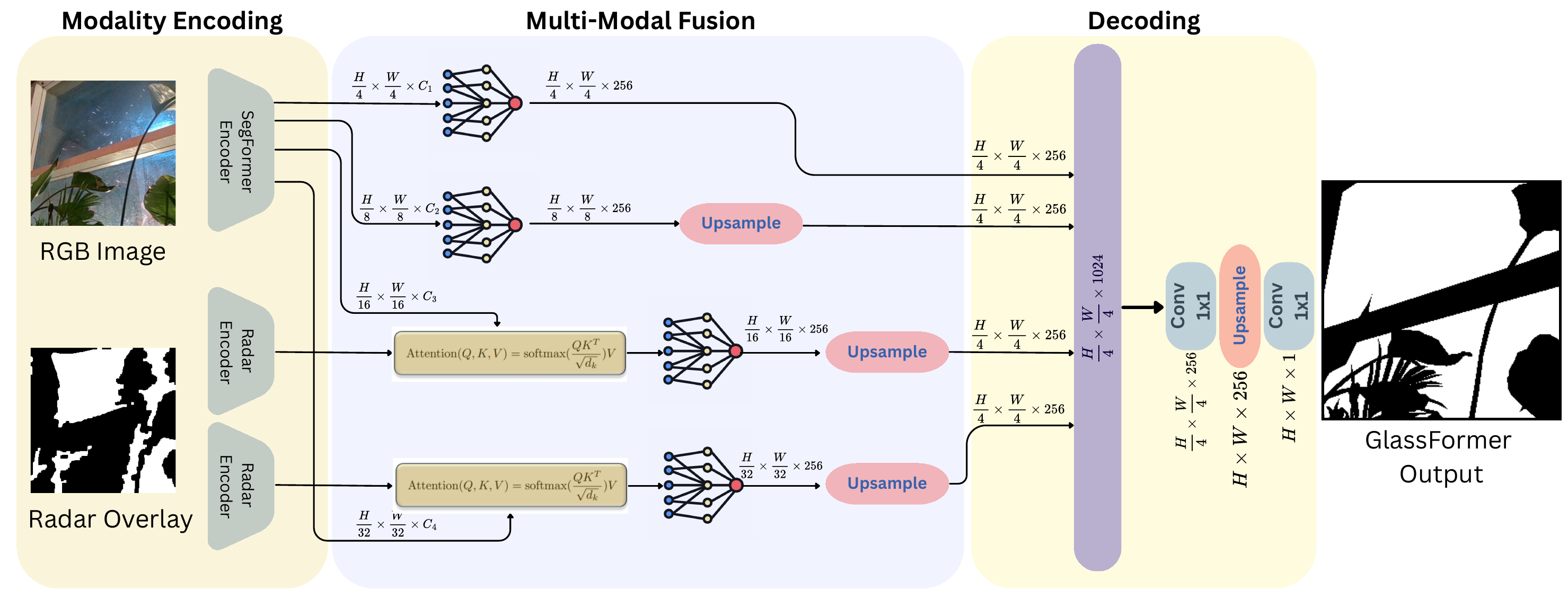}
  \caption{GlassFormer architecture. A SegFormer-B2 encoder extracts a
    4-scale RGB feature pyramid ($H/4$ to $H/32$). At the two
    lowest-resolution stages, a 2-layer CNN encodes the radar mask into
    queries, while RGB features serve as keys and values in a
    RadarAttention (RA) module; a learnable spatial gate $\lambda$
    controls how strongly the radar prior modulates each RGB feature,
    letting the model down-weight radar noise. The two RA-modulated
    stages and the two unmodified high-resolution stages are
    concatenated and decoded by an MLP followed by two $1\times1$
    convolutions to produce the segmentation mask.}
  \label{fig:methodology-flowchart}
\end{figure*}
An example is shown in Fig.~\ref{fig:radar_mask_generation}. The resulting binary mask is refined using morphological opening and closing with a $5 \times 5$ kernel to suppress isolated artifacts and fill small discontinuities, yielding a per-frame transparent-surface prior for the segmentation network described in Section~\ref{sec:model}.

\subsection{GlassFormer Architecture}
\label{sec:model}
Our model architecture, GlassFormer, builds on SegFormer-B2~\cite{segformer} with lightweight RadarAttention (RA) modules at the deeper encoder stages (Fig. \ref{fig:methodology-flowchart}). 

\subsubsection{SegFormer Backbone}
SegFormer consists of a hierarchical Mix Transformer (MiT) encoder paired with a lightweight MLP decoder. The encoder processes an input RGB image through four stages of overlapping patch merging and efficient self-attention, while producing multi-scale feature maps at resolutions $\frac{H}{4}, \frac{H}{8}, \frac{H}{16}, \frac{H}{32}$ with channel dimensions $\{64,\,128,\,320,\,512\}$ for the B2 variant.

The MiT encoder has a Mix-FFN, which embeds a $3{\times}3$ depth-wise convolution inside each feed-forward block to give positional information implicitly. This removes the need for fixed positional encodings that degrade when test resolution differs from training resolution. Efficient self-attention further reduces the sequence length of keys and values by a reduction ratio across stages, cutting computational complexity from $\mathcal{O}(N^2)$ to $\mathcal{O}(N^2/R)$ and enabling operation at high spatial resolutions. 
\begin{figure*}[t]
    \centering
    \setlength{\tabcolsep}{2pt}
    \renewcommand{\arraystretch}{1.0}
    \begin{tabular}{ccccccc}

        \includegraphics[width=0.13\textwidth,height=2.6cm]{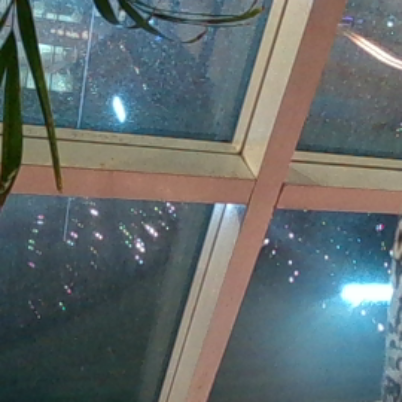} &
        \includegraphics[width=0.13\textwidth,height=2.6cm]{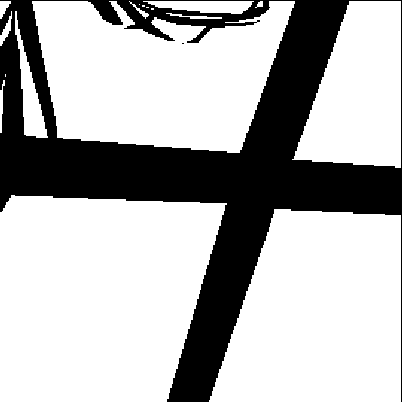}  &
        \includegraphics[width=0.13\textwidth,height=2.6cm]{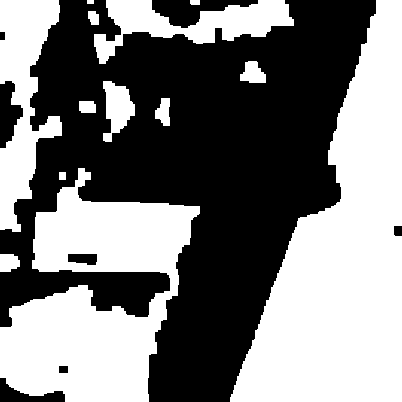}  &
        \includegraphics[width=0.13\textwidth,height=2.6cm]{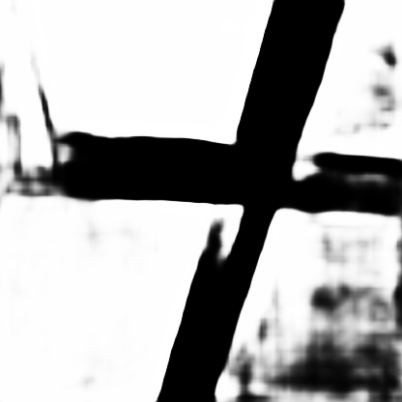}  &
        \includegraphics[width=0.13\textwidth,height=2.6cm]{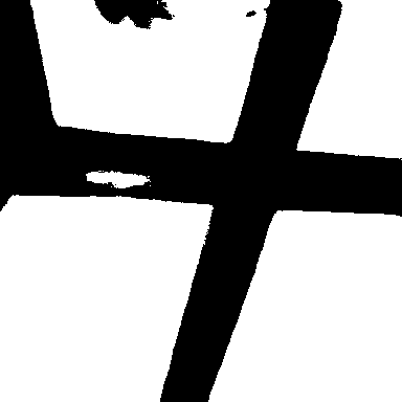}  &
        \includegraphics[width=0.13\textwidth,height=2.6cm]{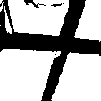}  &
        \includegraphics[width=0.13\textwidth,height=2.6cm]{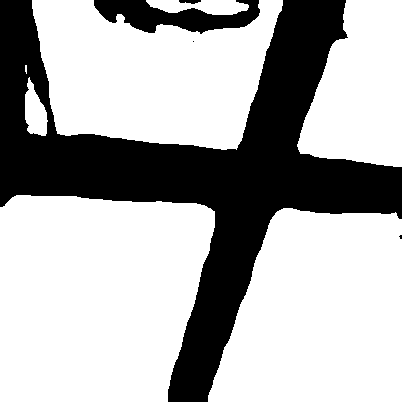}  \\

        \includegraphics[width=0.13\textwidth,height=2.6cm]{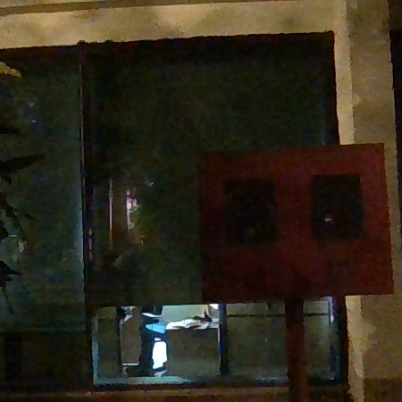} &
        \includegraphics[width=0.13\textwidth,height=2.6cm]{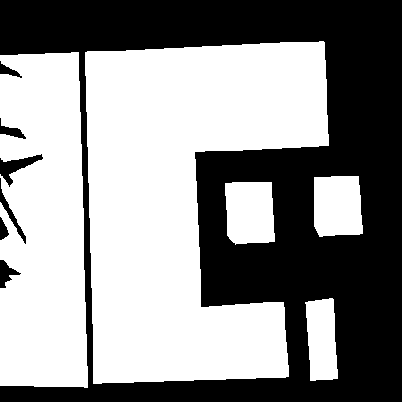}  &
        \includegraphics[width=0.13\textwidth,height=2.6cm]{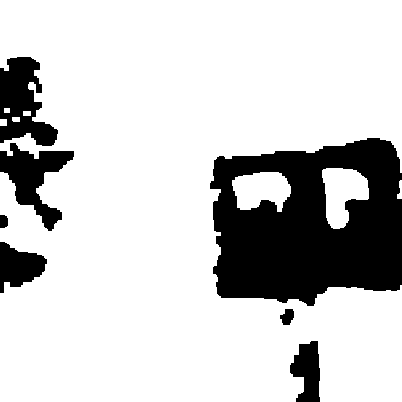}  &
        \includegraphics[width=0.13\textwidth,height=2.6cm]{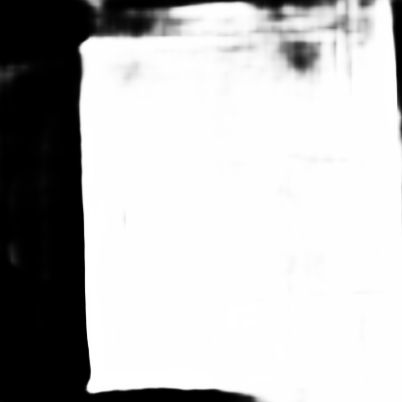}  &
        \includegraphics[width=0.13\textwidth,height=2.6cm]{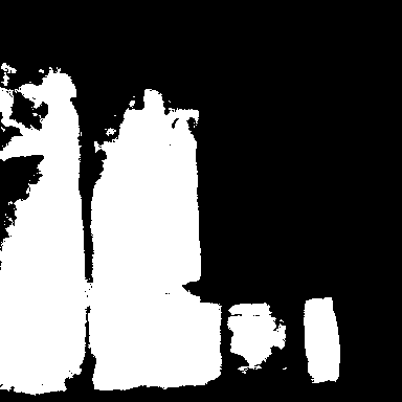}  &
        \includegraphics[width=0.13\textwidth,height=2.6cm]{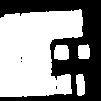}  &
        \includegraphics[width=0.13\textwidth,height=2.6cm]{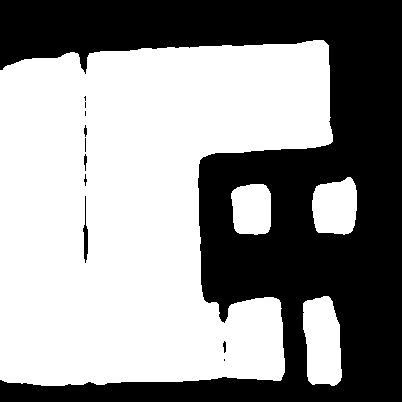}  \\

        \includegraphics[width=0.13\textwidth,height=2.6cm]{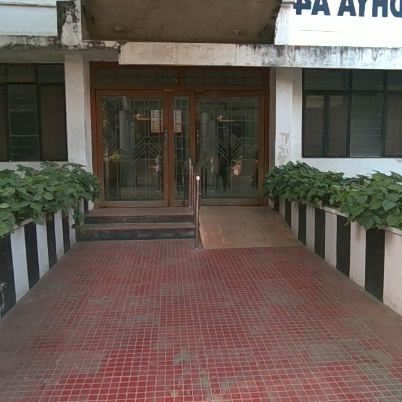} &
        \includegraphics[width=0.13\textwidth,height=2.6cm]{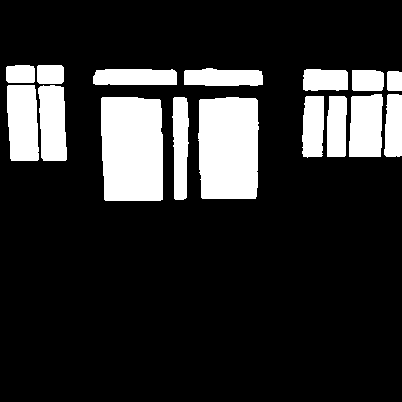}  &
        \includegraphics[width=0.13\textwidth,height=2.6cm]{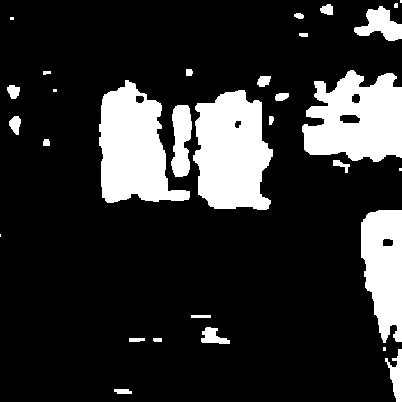}  &
        \includegraphics[width=0.13\textwidth,height=2.6cm]{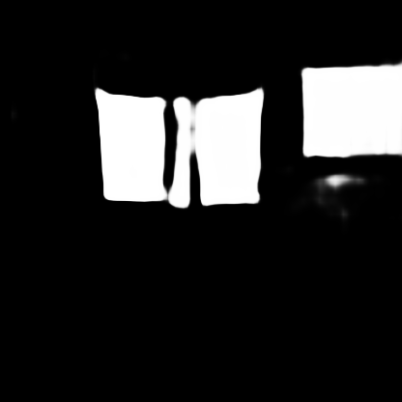}  &
        \includegraphics[width=0.13\textwidth,height=2.6cm]{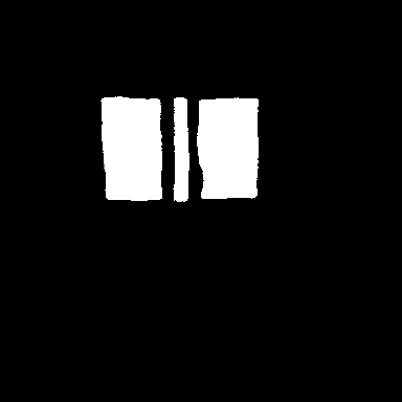}  &
        \includegraphics[width=0.13\textwidth,height=2.6cm]{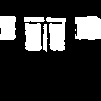}  &
        \includegraphics[width=0.13\textwidth,height=2.6cm]{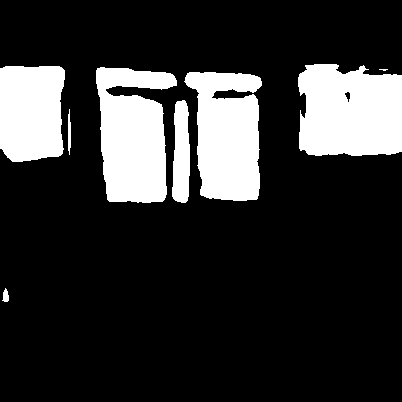}  \\

        \includegraphics[width=0.13\textwidth,height=2.6cm]{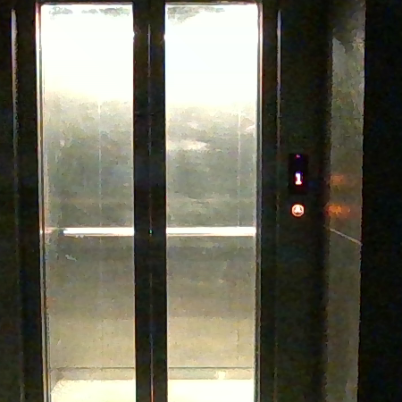} &
        \includegraphics[width=0.13\textwidth,height=2.6cm]{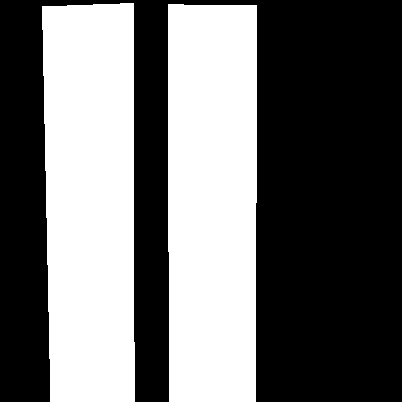}  &
        \includegraphics[width=0.13\textwidth,height=2.6cm]{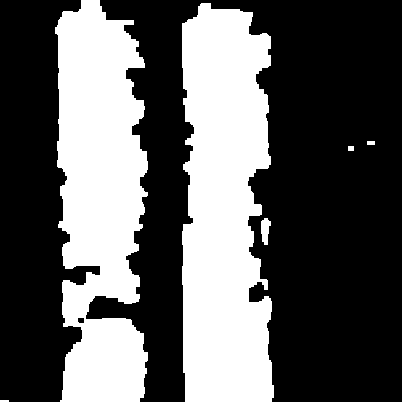}  &
        \includegraphics[width=0.13\textwidth,height=2.6cm]{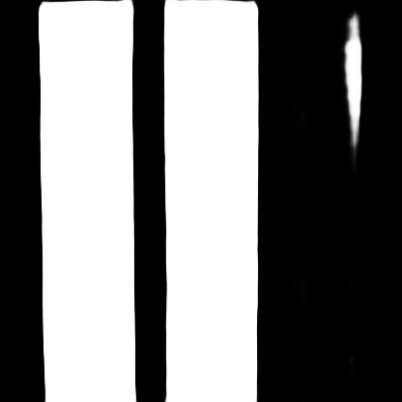}  &
        \includegraphics[width=0.13\textwidth,height=2.6cm]{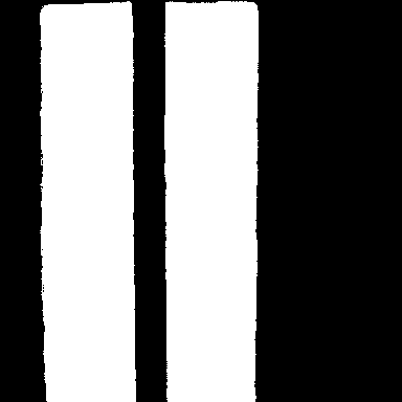}  &
        \includegraphics[width=0.13\textwidth,height=2.6cm]{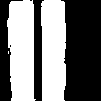}  &
        \includegraphics[width=0.13\textwidth,height=2.6cm]{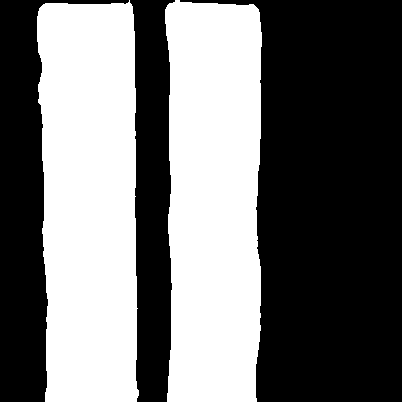}  \\

        \includegraphics[width=0.13\textwidth,height=2.6cm]{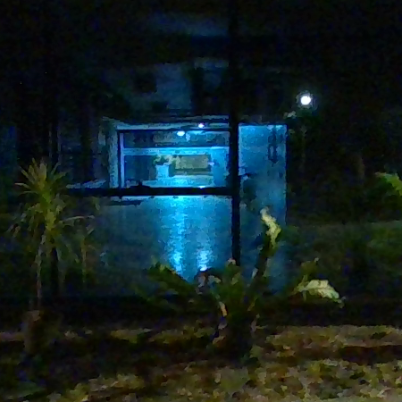} &
        \includegraphics[width=0.13\textwidth,height=2.6cm]{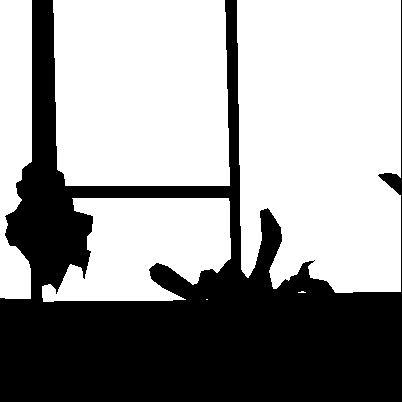}  &
        \includegraphics[width=0.13\textwidth,height=2.6cm]{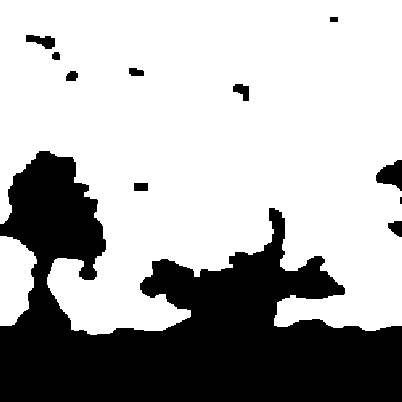}  &
        \includegraphics[width=0.13\textwidth,height=2.6cm]{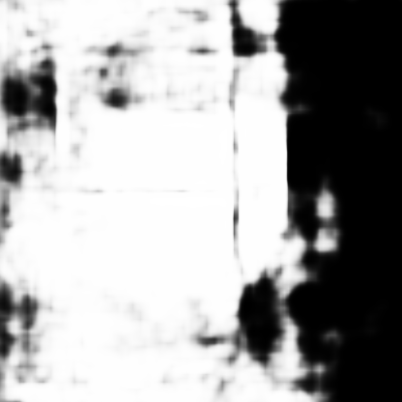}  &
        \includegraphics[width=0.13\textwidth,height=2.6cm]{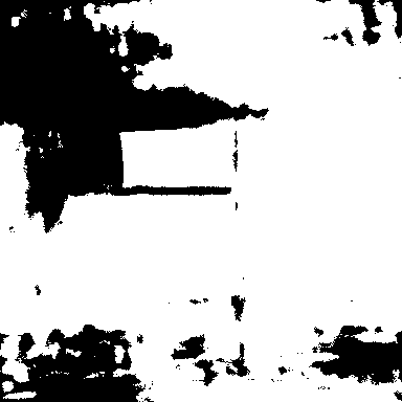}  &
        \includegraphics[width=0.13\textwidth,height=2.6cm]{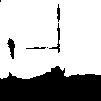}  &
        \includegraphics[width=0.13\textwidth,height=2.6cm]{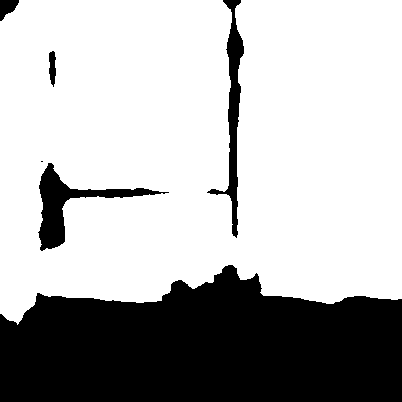}  \\

        \includegraphics[width=0.13\textwidth,height=2.6cm]{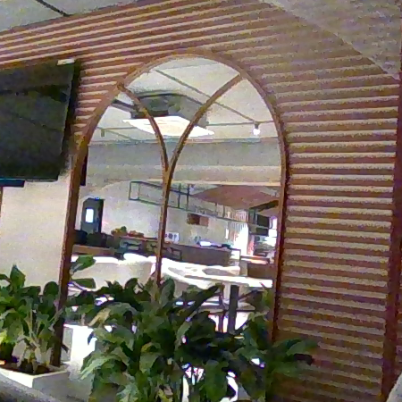} &
        \includegraphics[width=0.13\textwidth,height=2.6cm]{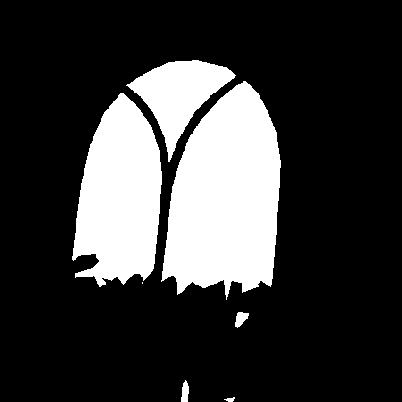}  &
        \includegraphics[width=0.13\textwidth,height=2.6cm]{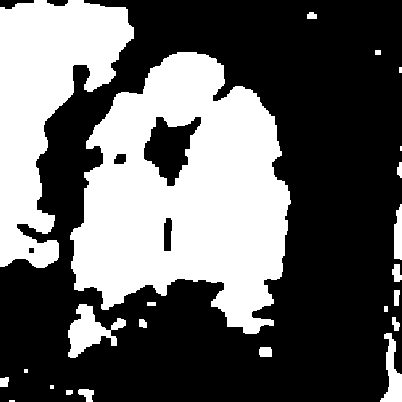}  &
        \includegraphics[width=0.13\textwidth,height=2.6cm]{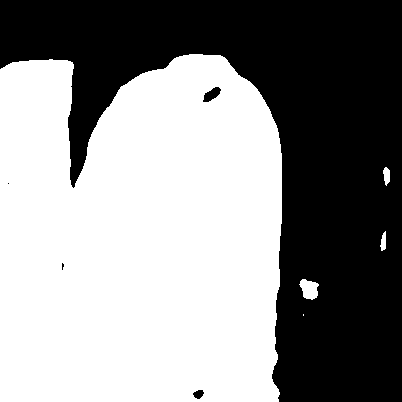}  &
        \includegraphics[width=0.13\textwidth,height=2.6cm]{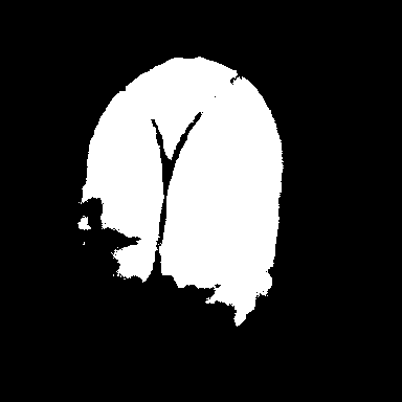}  &
        \includegraphics[width=0.13\textwidth,height=2.6cm]{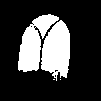}  &
        \includegraphics[width=0.13\textwidth,height=2.6cm]{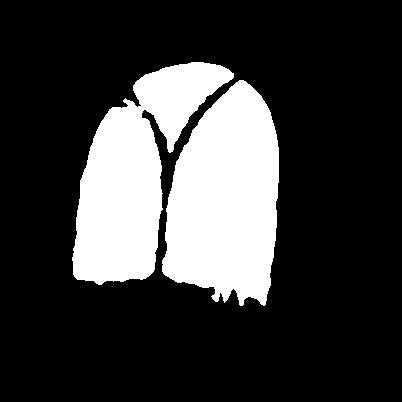}  \\

        \scriptsize RGB &
        \scriptsize Ground Truth &
        \scriptsize Radar Overlay &
        \scriptsize GDNet \cite{gdnet} &
        \scriptsize GlassSemNet \cite{glasssemnet} &
        \scriptsize SegFormer \cite{segformer} &
        \scriptsize Ours (GlassFormer)

    \end{tabular}
    \caption{Qualitative comparison across scenes (top to bottom: window, occluded glass wall night outdoor, glass door, elevator glass in low light, night outdoor, mirror) against baselines.}
    \label{fig:sota_qualitative}
\end{figure*}
\begin{table*}[h]
\centering
\caption{Comparison with state-of-the-art glass segmentation methods in well-lit conditions}
\label{tab:sota_comparison}
\setlength{\tabcolsep}{8pt}
\renewcommand{\arraystretch}{1.2}
\begin{tabular}{lcccc}
\toprule
\textbf{Method} 
& \textbf{mIoU} $\uparrow$ 
& \textbf{MAE} $\downarrow$ 
& \textbf{F-measure} $\uparrow$ 
& \textbf{BER} $\downarrow$ \\
\midrule

GDNet \cite{gdnet} & 0.5485 & 0.2894 & 0.6951 & 0.2618 \\
GlassSemNet \cite{glasssemnet} & 0.6336 & 0.1945 & 0.7769 & -- \\
SegFormer \cite{segformer} & 0.8799 & 0.0938 & 0.9361 & 0.0517 \\
Radar Overlay & 0.4687 & 0.3474 & 0.6382 & 0.3348 \\
\midrule
\textbf{Ours (GlassFormer)} & 
\textbf{0.8818} & 
\textbf{0.0835} & 
\textbf{0.9372} & 
\textbf{0.0517} \\
\bottomrule
\end{tabular}
\end{table*}

\begin{table*}[h]
\centering
\caption{Comparison with state-of-the-art glass segmentation methods in low light conditions}
\label{tab:sota_comparison_lowlight}
\setlength{\tabcolsep}{8pt}
\renewcommand{\arraystretch}{1.2}
\begin{tabular}{lcccc}
\toprule
\textbf{Method} 
& \textbf{mIoU} $\uparrow$ 
& \textbf{MAE} $\downarrow$ 
& \textbf{F-measure} $\uparrow$ 
& \textbf{BER} $\downarrow$ \\
\midrule

GDNet & 0.4824 & 0.4031 & 0.682 & 0.3981 \\
GlassSemNet & 0.4623 & 0.3620 & 0.7010 & -- \\
SegFormer & 0.4912 & 0.3588 & 0.6588 & 0.2819 \\
Radar Overlay & 0.5407 & 0.3148 & 0.7019 & 0.3097 \\
\midrule
\textbf{Ours (GlassFormer)} & 
\textbf{0.5904} & 
\textbf{0.2895} & 
\textbf{0.7424} & 
\textbf{0.2406} \\
\bottomrule
\end{tabular}
\end{table*}

The MLP decoder unifies the channel dimension across all four stages. Then, the features are upsampled to $\frac{H}{4}$ and concatenated together, after which an MLP layer is adopted to fuse these concatenated features. Finally, another MLP layer takes the fused features to predict a segmentation mask with $\frac{H}{4} \times \frac{W}{4} \times N_{\mathrm{cls}}$. This decoder design takes advantage of the large effective receptive field of the transformer encoder, which eliminates the need for heavy context modules (e.g.\ ASPP~\cite{aspp}) required by CNN-based decoders.

\subsubsection{RadarAttention Module}
The last two MiT stages operate at resolutions ($\frac{H}{16} \times \frac{W}{16}$ and
$\frac{H}{32} \times \frac{W}{32}$),
where features are semantically rich but spatially compressed.
We insert a RadarAttention (RA) module after each of these stages to incorporate the radar-derived transparency prior at the feature level.
Stages 1 and 2 are left unmodified since the cost of computing the attention maps at those resolution outweighs any advantage that the radar might give.

Each RA module takes a backbone feature map and the radar mask, downsamples to $(h, w)$ via bilinear interpolation, and passes through a two-layer convolutional encoder to produce a dense embedding at the same spatial resolution as the feature map.
Backbone features from RGB serve as key and value, while radar embeddings provide queries. A learnable spatial gating $\lambda$ is also used to control how much influence attention maps have on the RGB features. This allows the model to learn to reject false positives that radar might introduce due to noise in the depth estimates and resolution limitation. 
\begin{equation}
    Q = W_Q E_M, \quad K = W_K F, \quad V = W_V F,
\end{equation}
\begin{equation}
    F' = \mathrm{Softmax}\!\left(\frac{QK^\top}{\sqrt{d_k}}\right)V \times \lambda + F,
\end{equation}
where $d_k$ is the key dimension, $E_M$ is radar embedding and $F$ is the RGB embedding.

\subsubsection{Loss Function}
To supervise the segmentation output, we employ a combination of binary cross-entropy (BCE) loss and Lovász hinge loss. The BCE loss $\ell_{\mathrm{bce}}$ operates at the pixel level and provides stable gradients during training by penalizing misclassified pixels. The Lovász formulation directly optimizes the intersection-over-union objective by sorting prediction errors and computing the gradient of the Lovász extension of the Jaccard loss. This property makes it particularly effective for segmentation tasks with class imbalance or thin structures, such as transparent glass surfaces.

The final training objective combines the two losses with equal weight:
\begin{equation}
    \mathcal{L} = \ell_{\mathrm{bce}} + \ell_{\mathrm{lovasz}}.
\end{equation}
This combination encourages both accurate pixel-wise classification and improved region-level overlap, leading to more stable training and better alignment with the evaluation metrics used in our experiments.
\section{Experiments and Results}
\subsection{Implementation Details}
All models were implemented in PyTorch and trained on an NVIDIA RTX-4060 laptop GPU with 8 GB VRAM and 16 GB RAM. Inference runs at the speed of 70.6 fps, ensuring real-time performance. The model can also be inferred at 13.2 fps on a computer with Ryzen 9 6900hs CPU with 16 GB RAM. 
\subsection{Dataset}
The Acconeer XM125 pulsed coherent radar is aligned with the Intel RealSense D455 using a custom 3D printed mount (shown in Fig.~\ref{fig:range_profile}). The CAD model of the mount is used to obtain the static transform between the radar and RGB-D camera sensor axes. We tested our proposed method on a self-collected dataset of  1800 synchronised RGB-D-radar frames across diverse indoor and outdoor scenes. The dataset includes common cases of glass doors, windows, and full-length glass walls as well as mirrors, tinted glass, and partially occluded glass (as shown in Fig. \ref{fig:sota_qualitative}). 
Though the mirror instances are less numerous as compared to transparent glass cases, given their relative scarcity in the environments sampled. 

The data sequences are captured under a range of lighting conditions, distances, and angles. Cross-domain evaluation on existing glass segmentation benchmarks (e.g., Trans10K, GDD) was not performed, as these datasets do not include synchronized radar data required by our proposed method.
\begin{figure}[H]
  \centering  \includegraphics[width=0.9\columnwidth]{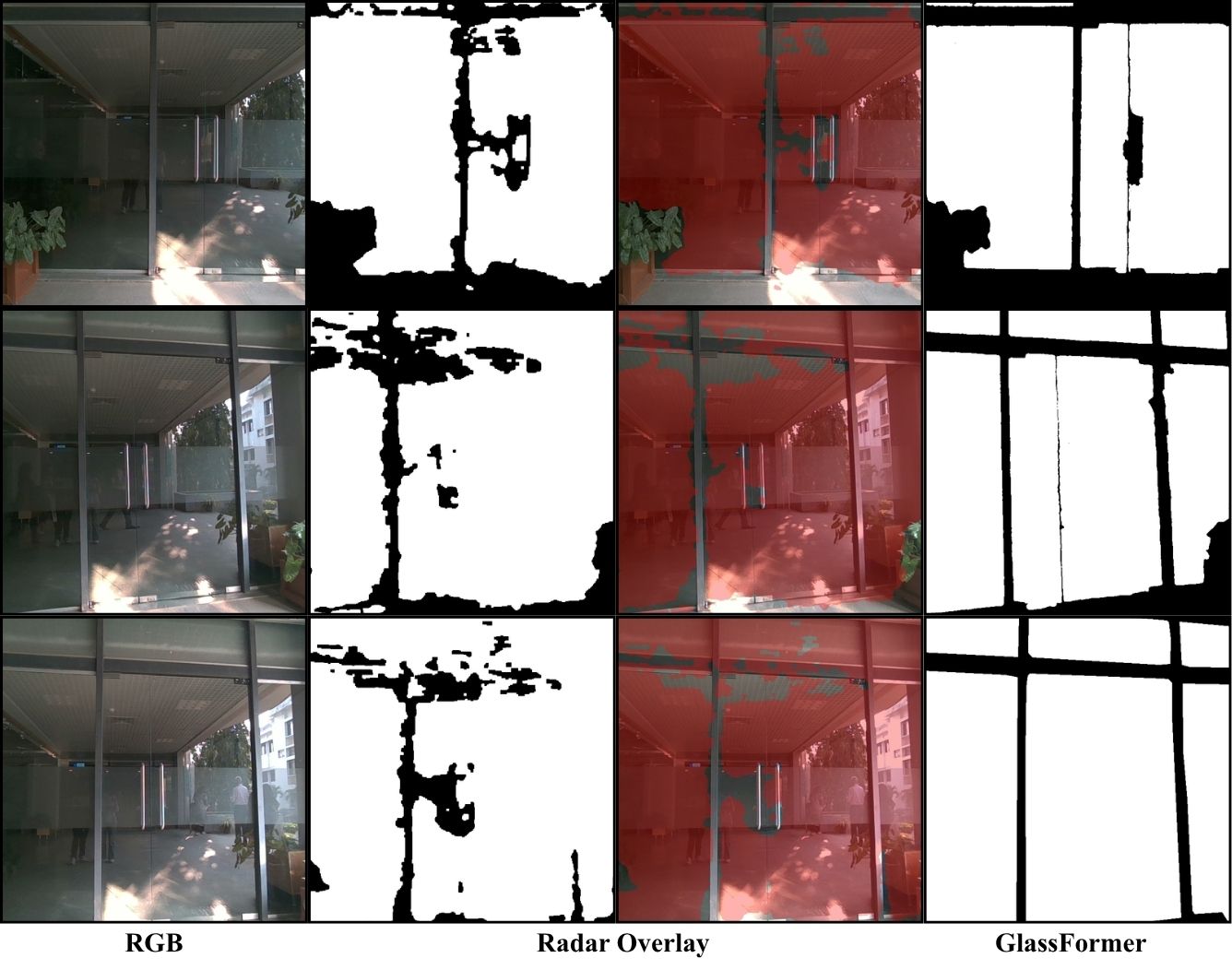}  \caption{GlassFormer across viewing angles and distances. Segmentation stays reliable across the radar–RGB-D shared FoV, not just frontal viewing} \label{fig:range_angle}
\end{figure}

\subsection{Effect of Radar Prior}
We first evaluate the standalone radar-derived binary mask (Radar Overlay) against the learned baselines. As shown in Tables~\ref{tab:sota_comparison} and~\ref{tab:sota_comparison_lowlight}, the radar-only mask underperforms vision-based methods in well-lit conditions (0.469 mIoU) but is competitive with, and in low light exceeds, several vision-only baselines (0.541 vs. 0.462–0.491 mIoU for GDNet, GlassSemNet, and SegFormer). This reflects complementary failure modes, that is, the radar-derived prior depends on depth-radar geometric consistency rather than visual appearance, so it remains stable regardless of illumination, but is susceptible to radar multipath/interference and to RGB-D depth artifacts under glare or strong reflections. Vision-only methods perform best when lighting and texture are favorable but degrade sharply once these cues disappear, motivating fusion rather than reliance on either modality alone.

\subsection{Effect of RadarAttention}
Incorporating the RadarAttention (RA) yields a modest gain over the fine-tuned SegFormer backbone in well-lit conditions (0.882 vs. 0.880 mIoU), where RGB cues are already near-saturated, but a substantially larger gain in low light (0.590 vs. 0.491 mIoU, BER improves from 0.282 to 0.241). This supports our hypothesis that the radar-guided prior contributes most precisely where RGB features are weakest, letting RadarAttention selectively redirect the model toward candidate transparent regions when visual evidence alone is insufficient, rather than uniformly biasing predictions irrespective of condition.

\section{Discussion and Conclusion}
Transparent surfaces present a fundamental challenge for robotic perception due to the limitations of optical sensing. While transformer-based methods demonstrate strong robustness even under challenging lighting conditions, they remain vulnerable to structural ambiguity, glare, and depth sensing failures.

In this work, we demonstrate that millimeter-wave radar provides a complementary geometric signal that can be integrated as a spatial prior within a transformer architecture. Although radar alone is insufficient for dense segmentation, its fusion via RadarAttention consistently improves region overlap and boundary refinement without degrading classification stability. Fig.~\ref{fig:sota_qualitative} (last row) shows a representative mirror scene, where GlassFormer is able to segment the reflective surface despite the smaller number of mirror examples in the training set. In particular, the proposed approach improves performance in low-light scenes. Importantly, the proposed approach maintains real-time performance and requires only lightweight modifications to an existing segmentation backbone, making it suitable for deployment on resource-constrained robotic platforms. 

However, the use of a single 1D radar introduces inherent spatial sparsity and can produce noisy reflections, which limits precise localization. Similarly, depth-guided masking strategies often misinterpret transparent surfaces as free space due to background depth returns, leading to systematic segmentation errors. By incorporating radar as a geometric prior within the transformer attention mechanism, our method mitigates these limitations by guiding the model toward regions where optical cues alone are unreliable while preserving the strong contextual reasoning of the vision backbone. Finally, while the radar itself supports ranging up to \SI{13}{\metre}, the effective operating range of the current sensor setup is limited to approximately \SI{6}{\metre} by the depth range of the Intel RealSense D455 used for depth-consistency filtering. This is a sensor-integration limitation rather than a limitation of the radar-guided approach; pairing the radar with a longer-range depth sensor would extend the effective segmentation range which we note as a direction for extending safe-navigation range in future deployments.

Future work will extend the proposed framework by exploiting the complex in-phase and quadrature (IQ) radar signal, including phase information rather than only amplitude and range, which may improve sub-wavelength surface discrimination and reduce ambiguity in the depth-consistency prior;enabling richer geometric reasoning and more accurate transparent surface segmentation. We also plan to explore uncertainty-aware fusion, dynamic radar confidence weighting, and broader evaluation under extreme illumination degradation and adverse weather conditions.

\section{Acknowledgement}
We acknowledge IHub-Data for supporting this work via project
M2-029.




\bibliographystyle{IEEEtran}
\bibliography{references}

\end{document}